\documentclass{IEEEtran}
\usepackage{cite}
\usepackage{amsmath,amssymb,amsfonts}
\usepackage{algorithmic}
\usepackage{graphicx,color}
\usepackage{textcomp}
\usepackage{xcolor}
\usepackage{hyperref}
\usepackage{verbatim}

\usepackage[utf8]{inputenc}
\usepackage[T1]{fontenc}
\usepackage{authblk}
\usepackage[most]{tcolorbox}
\usepackage{inconsolata} 

\definecolor{genbg}{HTML}{F0F8FF}     
\definecolor{genframe}{HTML}{2B547E}  
\definecolor{valbg}{HTML}{F5FFFA}     
\definecolor{valframe}{HTML}{2E8B57}  
\definecolor{promptbg}{HTML}{FFFFFF}  
\definecolor{promptframe}{HTML}{808080} 

\newtcolorbox{promptbox}[1]{
  enhanced,
  breakable,
  colback=promptbg,
  colframe=promptframe,
  coltitle=black,
  fonttitle=\bfseries\sffamily,
  title=#1,
  boxrule=0.5pt,
  attach boxed title to top left={yshift=-2.5mm, xshift=4mm},
  boxed title style={colback=promptbg, colframe=promptframe, size=small},
  top=4mm,
  fontupper=\ttfamily\small, 
  left=3mm, right=3mm, bottom=3mm
}

\hypersetup{hidelinks=true}
\usepackage{algorithm,algorithmic}
\def\BibTeX{{\rm B\kern-.05em{\sc i\kern-.025em b}\kern-.08em
    T\kern-.1667em\lower.7ex\hbox{E}\kern-.125emX}}
\AtBeginDocument{\definecolor{tmlcncolor}{cmyk}{0.93,0.59,0.15,0.02}\definecolor{NavyBlue}{RGB}{0,86,125}}
\usepackage{multirow}

\usepackage{graphicx}
\usepackage{subcaption}  
\usepackage{float} 
\newcommand{\llm}{\texttt{LLM-AUG} }

\begin{document}

\markboth{LLM-AUG}{Gajjar {et al.}}

\title{\texttt{LLM-AUG}: Robust Wireless Data Augmentation with In-Context Learning in Large Language Models}
\author{Pranshav Gajjar, Manan Tiwari, Sayanta Seth, and Vijay K. Shah}
\affil{\textit{NextG Wireless Lab}, North Carolina State University, NC}

\maketitle

\begin{abstract}
Data scarcity remains a fundamental bottleneck in applying deep learning to wireless communication problems, particularly in scenarios where collecting labeled Radio Frequency (RF) data is expensive, time-consuming, or operationally constrained. This paper proposes \texttt{\textbf{LLM-AUG}}, a data augmentation framework that leverages in-context learning in large language models (LLMs) to generate synthetic training samples directly in a learned embedding space. Unlike conventional generative approaches that require training task-specific models, \texttt{LLM-AUG} performs data generation through structured prompting, enabling rapid adaptation in low-shot regimes. We evaluate \texttt{LLM-AUG} on two representative tasks: modulation classification and interference classification using the RadioML 2016.10A dataset, and the Interference Classification (IC) dataset respectively. Results show that \texttt{LLM-AUG} consistently outperforms traditional augmentation and deep generative baselines across low-shot settings and reaches near \textit{oracle} performance using \textit{only} \textasciitilde 15\% labeled data. \texttt{LLM-AUG} further demonstrates improved robustness under distribution shifts, yielding a 29.4\% relative gain over diffusion-based augmentation at a lower SNR value. On the RadioML and IC datasets, \llm yields a relative gain of 67.6\% and 35.7\% over diffusion-based baseline. The t-SNE visualizations further validate that synthetic samples generated by \texttt{LLM-AUG} better preserve class structure in the embedding space, leading to more consistent and informative augmentations. These results demonstrate that LLMs can serve as effective and practical data augmenters for wireless machine learning, enabling robust and data-efficient learning in evolving wireless environments.
\end{abstract}


\section{Introduction}
\label{intro}

Machine learning is increasingly used in wireless systems for tasks such as modulation classification, interference identification, and intelligent spectrum management \cite{chen2019artificial, huynh2021automatic}. However, the performance of these methods depends heavily on access to labeled Radio Frequency (RF) data, which is often difficult and expensive to collect. Unlike computer vision or natural language processing, wireless dataset construction requires specialized hardware, controlled experiments, and reliable signal labels, while operational datasets are often restricted by privacy, security, or proprietary constraints \cite{soltani2024learning, li2022unsupervised}. As a result, many wireless learning problems must be addressed in low-data settings.
This limitation is particularly problematic because wireless models are expected to generalize across diverse signal-to-noise ratios (SNRs), channel conditions, hardware platforms, and interference environments. When only a few labeled samples are available per class, classifiers often overfit the observed data and fail to generalize to unseen operating conditions. Improving learning performance in such low-shot regimes remains a central challenge in wireless machine learning.


More broadly, learning-driven wireless systems face four key challenges: (i) scarcity of labeled RF data, (ii) model drift due to dynamically evolving wireless environments, (iii) concept drift arising from changing interference patterns and network conditions, and (iv) limited generalization in low-shot regimes. Addressing these challenges requires data-efficient methods that can expand the effective training distribution without relying on extensive data collection or frequent retraining. \textit{Data augmentation} is a natural way to address this problem by expanding the effective training set with synthetically generated samples. In wireless communications, data augmentation has traditionally relied on domain-informed transformations such as noise injection, phase rotation, frequency shifting, and time scaling \cite{oshea2016radio}. While simple and label preserving, these methods usually generate only small perturbations of existing samples and often fail to capture the broader variability of real wireless environments. Deep generative models such as Generative Adversarial Networks (GANs), Variational Auto Encoders (VAEs), and diffusion models offer a more expressive alternative (see the detailed discussion in Section II), but they typically require larger datasets and can become unstable in extremely low-shot settings, where overfitting, memorization, and mode collapse are common \cite{cao2023surveygenerativediffusionmodel}. Recent advances in large language models (LLMs) suggest a different approach. Rather than training a dedicated generator, an LLM can use in-context learning to infer structure from a small set of examples and produce new samples without parameter updates. However, directly generating raw RF signals or spectrograms with an LLM is impractical because such representations are high-dimensional and inefficient to include in prompts. This motivates a more compact formulation in which augmentation is performed in a learned feature space instead of the raw signal domain.

In this work, we propose \textbf{\texttt{LLM-AUG}}, an embedding-space data augmentation framework for low-shot wireless learning. A pretrained convolutional neural network (CNN) first maps wireless spectrograms into compact embedding vectors that preserve class-relevant signal structure. A small labeled set of these embeddings is then provided to an LLM through a structured prompt, and the LLM generates additional synthetic embeddings using in-context learning. The synthetic and real embeddings are finally combined to train a downstream classifier. By operating in latent space, \texttt{LLM-AUG} avoids the difficulty of raw signal synthesis while eliminating the need to train a task-specific generative model. We evaluate \texttt{LLM-AUG} on representative wireless learning tasks, including automatic modulation classification and interference classification in O-RAN settings. The results show that the proposed approach consistently improves performance over both manual augmentation and deep generative baselines, particularly in low-data regimes. 
The key contributions of this work are summarized as follows:


\begin{itemize}[\leftmargin=3pt]
    \item 
    We propose a novel augmentation pipeline, \texttt{LLM-AUG}, that leverages in-context learning in large language models to generate synthetic feature embeddings directly in latent space, eliminating the need for training task-specific generative models.

    \item 
    We demonstrate that \llm is significantly more data-efficient than classical generative augmentation methods and reaches near \textit{oracle} performance with substantially fewer labeled samples, achieving this threshold at approximately 15\% of labeled data compared to about 21\% for VAE and 33\% for GAN.    

    \item 
    We experiment with two prominent datasets, RadioML \cite{oshea2016radio} and Interference Classification \cite{chiejina2024system}, and we observe how \llm consistently improves performance against multiple baselines with relative gains up to 67.6\% and 35.7\% for the diffusion-based baseline for RadioML and IC datasets, respectively.   

    \item 
    We further evaluate \textit{robustness} of our work by testing performance on unseen distributions, and \llm consistently outperforms diffusion-based methods with relative performance gains of around 29.4\%.

\end{itemize}

The remainder of this paper is organized as follows. Section II reviews background and related work on wireless data modalities and augmentation techniques. Section III presents the motivation and design rationale of \llm framework. Section IV describes the \llm framework in detail, along with the walkthrough of the pipeline. Section V outlines the experimental setup, followed by results and discussion in Sections VI and VII, respectively. Finally, Section VIII concludes the paper.

\section{Background and Related Work}
\label{sec:background}

This section reviews prior work on wireless signal representations and data augmentation strategies for learning under limited labeled data. We first summarize the signal representations and modeling paradigms commonly used in wireless machine learning, along with the fundamentals of LLMs and in-context learning. We then discuss existing augmentation approaches, including classical transformations, deep generative models, and emerging LLM-based methods.\\

\noindent \textbf{Background:} Wireless machine learning systems operate on signal representations derived from RF measurements, including in-phase and quadrature (IQ) time-series samples, power spectral density (PSD), constellation diagrams, and time-frequency representations such as spectrograms. Among these, spectrograms are widely used for tasks such as modulation classification and interference detection, as they enable the use of CNNs as de-facto ML model \cite{CNN,DNN,AIMC}. 

Unlike natural images, spectrograms encode structured signal characteristics determined by modulation formats, channel effects, and interference patterns. Consequently, transformations that are commonly used in computer vision do not necessarily preserve the physical semantics of the signal. In addition, spectrograms are typically high-dimensional representations, requiring the modeling of dependencies across both time and frequency domains. 

These challenges motivate the need for alternative approaches that can effectively capture structured relationships from limited data without explicitly modeling high-dimensional signal distributions. In this context, LLMs provide a flexible framework for learning from small sets of examples through in-context learning.
LLMs are probabilistic models based on the Transformer architecture that generate sequences in an autoregressive manner. Given a sequence of tokens $x_1, x_2, \dots, x_t$, the model predicts the probability of the next token by modeling $P(x_{t+1} | x_1, x_2, \dots, x_t)$, and constructs sequences by iteratively sampling from this distribution. The joint probability of a sequence of length $N$ is expressed as $P(x_1, \dots, x_N) = \prod_{i=1}^{N} P(x_i \mid x_1, \dots, x_{i-1})$. This formulation enables \textit{in-context learning}, where the model conditions on a prompt containing a task description and a small set of examples. If the prompt is denoted as $C$, the generation of a sequence $Y = (y_1, \dots, y_K)$ follows the conditional distribution $P(y_i \mid C, y_1, \dots, y_{i-1})$. By leveraging patterns learned during pretraining, the model can infer relationships within the provided context and generate consistent outputs without requiring parameter updates \cite{brown2020language, wei2022emergent}. This property enables rapid adaptation to new tasks in low-shot settings.\\

\noindent \textbf{Related Work:} Data augmentation is widely used to improve generalization by expanding the effective training dataset. In wireless learning, augmentation approaches generally fall into two categories. 

The first category consists of \textit{domain-informed transformations}, including noise injection, frequency shifting, and phase perturbation. These methods preserve label semantics and are straightforward to implement, but they typically generate only limited variations and remain close to the original data distribution. As a result, they often fail to capture the diversity observed in real-world wireless environments \cite{naveed2022assessing}. The second category includes \textit{data-driven generative models}, which learn to synthesize new samples directly from training data. Autoencoder-based approaches have been explored to introduce controlled variability while preserving underlying signal structure, demonstrating improvements in representation learning under moderate data availability \cite{AEWCL}. GANs, particularly conditional variants, have been applied to tasks such as automatic modulation classification to generate labeled synthetic samples that improve performance in supervised settings \cite{patel_GAN,pandey_GAN}. 
More recently, diffusion models have emerged as a class of probabilistic generative models that learn to reverse a process that gradually degrades the training data structure. Since 2020, diffusion-based approaches have become the state-of-the-art and have dominated visual generation and have driven recent progress in high-quality image synthesis \cite{ravishankar2026fairbenchmarkingemergingonestep}. These models consist of a forward process that progressively destroys data by adding noise and a reverse process that reconstructs the data by sequentially removing noise \cite{mubarak_diffusion}. At inference time, samples are generated by gradually reconstructing data starting from random noise through an iterative denoising procedure \cite{mubarak_diffusion, cao_diff}.
In wireless applications, diffusion models have been explored for data augmentation in tasks such as automatic modulation recognition, where they provide an alternative to GANs by removing the need for complex adversarial training, thereby enhancing model stability \cite{li_diffusion}. Due to their strong generation and denoising capabilities, these models have been increasingly applied in communication and signal processing domains. 
Despite their ability to generate diverse samples, these generative approaches rely on training data-driven models and therefore exhibit degraded performance in low-shot regimes. Limited data leads to overfitting, memorization, and insufficient coverage of the underlying signal distribution. In addition, these methods introduce significant computational overhead due to the need for training and maintaining task-specific generative architectures. Recent advances in LLMs introduce an alternative paradigm for data generation grounded in in-context learning. Rather than training a dedicated generative model, an LLM can infer underlying patterns from a small set of prompt-provided examples and produce new samples without any parameter updates \cite{brown2020language, wei2022emergent}. While this flexibility makes LLMs particularly attractive for low-shot scenarios, extending such approaches to wireless data remains challenging due to the high dimensionality and structured nature of signal representations. 

Taken together, existing augmentation strategies present a fundamental tradeoff in low-shot wireless learning. Classical transformations provide limited diversity, while generative models require sufficient data to learn reliable distributions. Although LLM-based approaches offer a promising alternative, their effectiveness is constrained by how wireless data is represented and presented to the model. These limitations motivate the need for an alternative formulation, which we develop in the next section.

\section{Motivation and Design Rationale}

This section analyzes the key limitations that hinder effective data augmentation in low-shot wireless learning and motivates the design of the proposed \texttt{LLM-AUG} framework. We focus on two fundamental challenges: (i) \textit{distributional drift in dynamic wireless environments}, which necessitates rapid adaptation, and (ii) the \textit{inefficiency of directly generating high-dimensional signal representations using LLMs}. Together, these challenges reveal the need for a compact and structured augmentation approach, leading to the embedding-space formulation introduced in this work.

\subsection{Data Drift and Concept Drift}
\label{subsec:model_drift}

\begin{figure}[htbp]
\centering
\includegraphics[width=0.8\columnwidth]{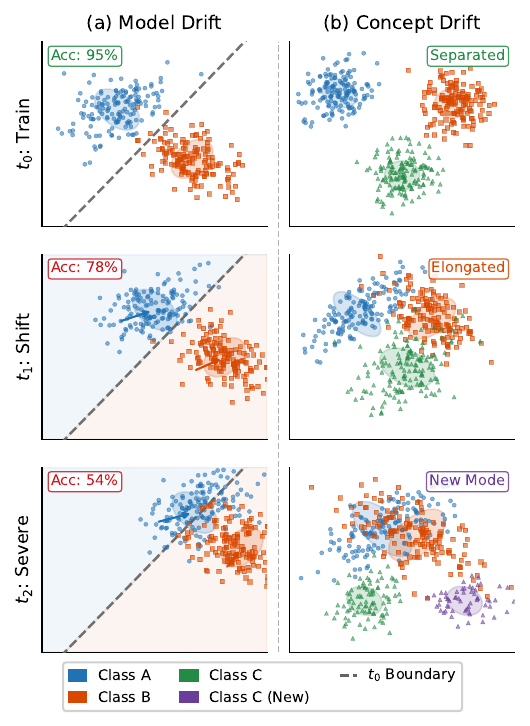}
\caption{Simulation of (a) model drift and (b) concept drift over three time steps using synthetic data. Model drift arises from translation under a fixed boundary, while concept drift includes structural changes and a new mode at $t_2$.}
\label{fig:drift}

\end{figure}

Dynamic wireless environments induce both data drift and concept drift, where the underlying signal distributions and class boundaries evolve over time. Fig. \ref{fig:drift} illustrates these effects. Class A and Class B represent baseline network states that shift across a fixed decision boundary, capturing model drift. In contrast, Class C undergoes structural change and splits into a new mode, representing concept drift due to emerging interference or environmental changes.

Under such conditions, models trained on historical data may become suboptimal as the deployment environment evolves \cite{liu2023leaf}. A common mitigation strategy is periodic retraining using newly collected data. However, retraining can be computationally expensive, operationally disruptive, and difficult to perform at the frequency required to track rapid environmental changes. Moreover, when only limited new data is available, retraining may lead to overfitting or unstable updates \cite{gudepu2024generative}.

These challenges highlight the need for lightweight adaptation mechanisms that can expand the effective training distribution without requiring full retraining. Data augmentation provides such a mechanism by introducing synthetic samples that approximate evolving conditions. In this context, augmentation methods that operate effectively in low-data regimes and can be deployed without additional model training are particularly desirable. This motivates the exploration of LLM-based augmentation as a fast and flexible alternative for adapting to both data and concept drift.

\subsection{In-context Learning with Spectrograms}

\begin{figure}[htbp]
\centering
\includegraphics[width=\columnwidth]{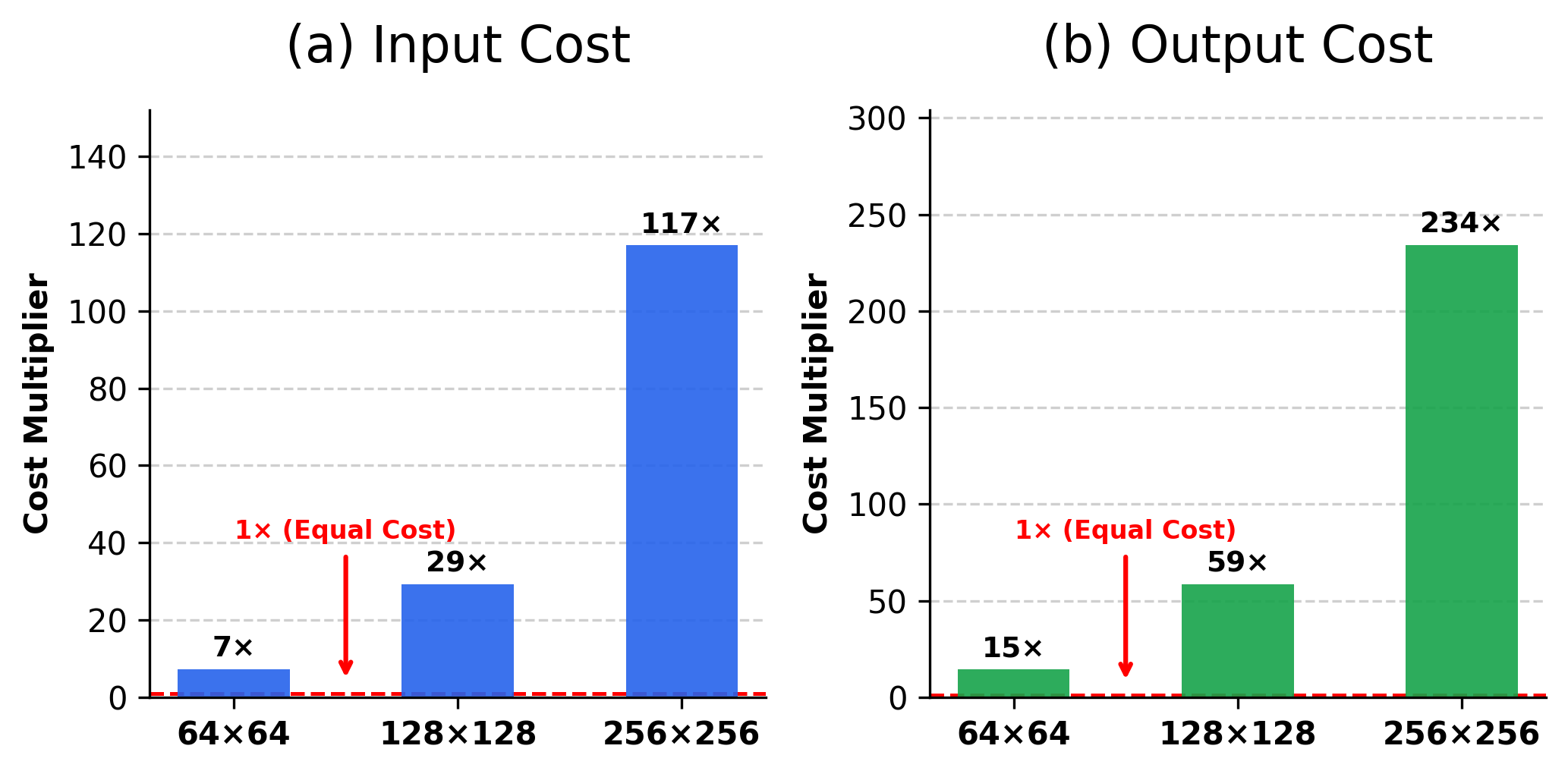}
\caption{Input and output costs pertaining to using vision for augmentation.}
\label{fig:cost}

\end{figure}

We examine whether LLMs can directly augment wireless spectrogram data. While providing labeled spectrograms to instruct a model to generate new ones seems intuitive, the in-context approach is limited in the low-shot wireless setting due to the dimensionality and structure of these representations. As illustrated in Fig. \ref{fig:cost}, directly modeling raw images through in-context prompting incurs prohibitive input and output computational costs compared to processing a compact 10-dimensional embedding vector, which serves as our $1\times$ equal cost baseline. Because vision models tokenize images based on pixel patches, both the token count and the associated financial costs scale quadratically with image resolution. This becomes a critical bottleneck given that standard spectrogram sizes across the wireless literature are typically $256 \times 256$, with some recent works utilizing even higher resolutions. At a $256 \times 256$ resolution, the input and output costs increase to $117\times$ and $234\times$ the baseline, respectively. Consequently, the prompt rapidly becomes dominated by image representation rather than task information, strictly limiting the number of examples that can be provided. Furthermore, these representations exhibit structured dependencies across time and frequency, which increases the complexity of direct generation. Further information regarding these preliminary results is available in Appendix \ref{appx:multimodality}. 

Therefore, our framework shifts the augmentation problem entirely to the embedding domain. By operating on compressed vectors, the LLM receives inputs that are both computationally efficient and semantically meaningful. Assuming the pretrained feature extractor produces geometrically coherent and discriminative embeddings, the LLM can infer the feature space geometry from a small labeled subset and generate plausible new embeddings. Guided by this motivation, the remainder of the paper formulates this latent space wireless augmentation system and evaluates its performance.

\section{LLM-AUG Framework}
\label{sec:sysmodel}
\begin{figure*}[h]
    \centering
    \includegraphics[width=1\linewidth]{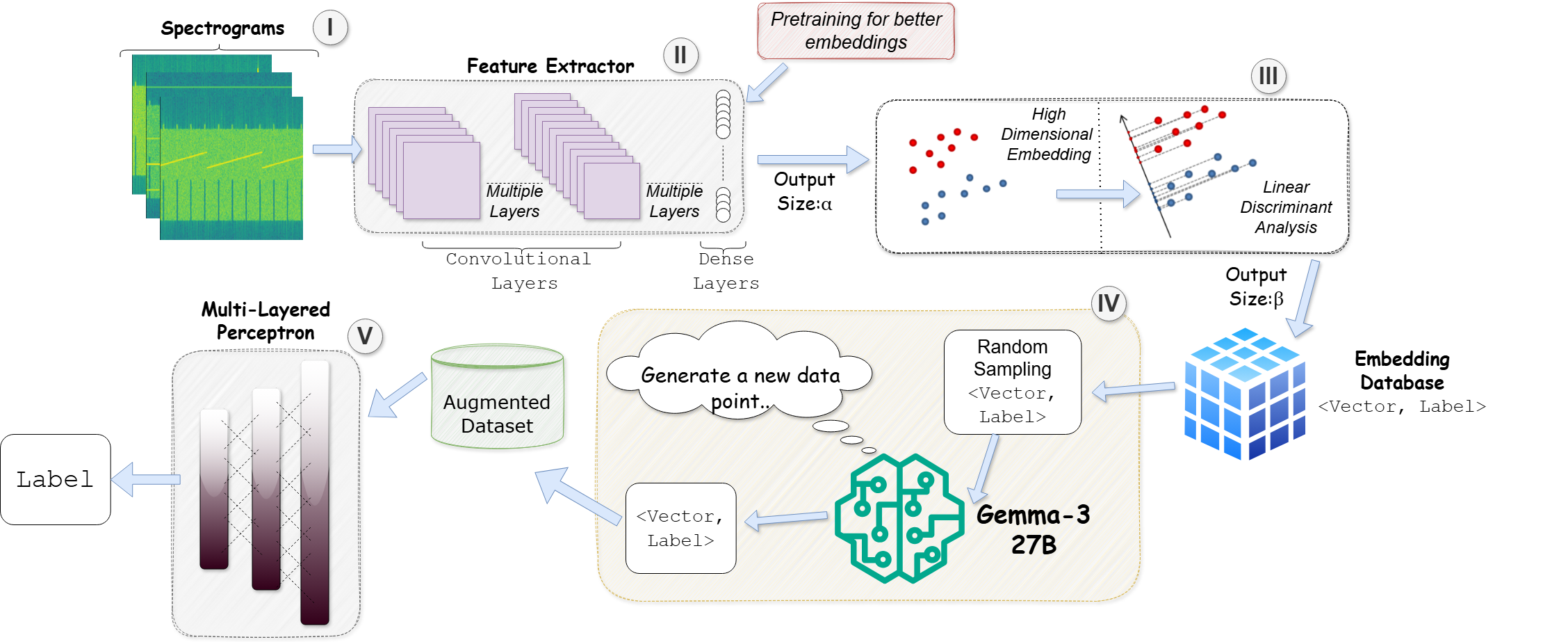}
    
    \caption{Overview of the proposed \texttt{LLM-AUG} framework.} 
    \label{fig:framework}
    
\end{figure*}
This section presents the proposed \llm framework. The method performs data augmentation directly in a learned embedding space by leveraging in-context generation with an LLM. Starting from spectrogram representations, signal samples are mapped to compact embeddings, which are then used to construct few-shot prompts for the LLM. The model generates additional synthetic embeddings that are combined with real samples to train a downstream classifier. 

\textit{End-to-end LLM-AUG workflow.} The overall workflow is illustrated in Fig.~\ref{fig:framework}. \textbf{(I)} Raw wireless signal observations are converted into time-frequency spectrograms, which serve as structured inputs to the feature extraction stage. \textbf{(II)} Each spectrogram is processed by a pretrained CNN that maps the input to a high-dimensional embedding vector $z \in \mathbb{R}^{\alpha}$.
\textbf{(III)} The extracted embeddings are optionally projected using Linear Discriminant Analysis to obtain a lower-dimensional representation $z' \in \mathbb{R}^{\beta}$ with improved class separability, where $\beta < \alpha$. \textbf{(IV)} The projected embeddings are stored in a labeled database. A small subset of embeddings is sampled to construct a few-shot prompt, which is provided to the LLM. The model generates additional synthetic embedding-label pairs that are consistent with the observed feature distribution.
, and \textbf{(V)} The generated embeddings are combined with the original dataset to form an augmented dataset, which is used to train an MLP classifier for signal classification.

In the remainder of the section, we describe  the core \llm components in detail.


\subsection{Feature Extractor}
\label{subsec:feature_extractor}
The feature extractor maps input spectrograms into a compact embedding space that captures the underlying structure of wireless patterns. Let $x \in \mathcal{X}$ denote a spectrogram representation derived from a raw wireless signal. A pretrained CNN encodes $x$ into a fixed-dimensional embedding vector $z = f_{\theta}(x)$, where $z \in \mathbb{R}^{\alpha}$. Here $f_{\theta}(\cdot)$ denotes the CNN parameterized by $\theta$, and $\alpha$ is the embedding dimensionality. The CNN consists of stacked convolutional layers followed by dense layers, which progressively transform local time-frequency patterns into higher-level representations. To establish this latent space, a CNN is pretrained using the available real samples, providing stable and semantically meaningful embeddings prior to the augmentation step. This design choice decouples representation learning from the downstream augmentation process and reduces the need for large labeled datasets. The resulting embeddings serve as the fundamental representation throughout the pipeline. All subsequent operations are performed in this embedding space. 
By operating on learned features instead of raw spectrograms, the framework ensures that augmentation is applied to structured and discriminative representations that capture class-specific characteristics of the signals.

\subsection{Embedding Projection}
\label{subsec:projection}
The embeddings produced by the feature extractor may lie in a high-dimensional space, which can make analysis and downstream processing less efficient. To obtain a more compact and discriminative representation, a projection step is applied using Linear Discriminant Analysis (LDA). Given the embedding vector $z\in \mathbb{R}^{\alpha}$ and its associated class label, LDA computes a linear transformation that maximizes class separability by balancing between-class and within-class scatter. The projected embedding is given by $z' = W^{\top} z, \quad z' \in \mathbb{R}^{\beta}$. Here $W \in \mathbb{R}^{\alpha \times \beta}$ is the projection matrix and $\beta$ denotes the reduced dimensionality. This transformation preserves the most discriminative directions in the embedding space while reducing dimensionality. The resulting projected embeddings $z'$ form the feature space used in the subsequent augmentation and classification stages. In addition, the lower-dimensional representation facilitates visualization and provides insight into class separability and the structure of both real and synthetic samples.

\subsection{LLM-Based Augmentation}
\label{subsec:llm_aug}

The augmentation stage operates directly in the embedding space by leveraging an LLM to generate additional feature vectors conditioned on a small set of labeled examples. 

Let $\mathcal{Z} = \{(z_i, y_i)\}_{i=1}^{N}$ denote the set of available embeddings, where $z_i \in \mathbb{R}^{\beta}$ is a projected embedding and $y_i$ is its corresponding class label. A subset of $k$ labeled embeddings is sampled to form a few-shot context set $S = \{(z_i, y_i)\}_{i=1}^{k}$, which is serialized into a structured prompt and provided to the LLM. Conditioned on $S$, the model generates new embedding-label pairs $(\tilde{z}, \tilde{y}) = g_{\text{LLM}}(S)$, where $g_{\text{LLM}}(\cdot)$ denotes the mapping induced by the LLM. By repeating this process, a collection of synthetic samples $\mathcal{Z}_{\text{synthetic}} = \{(\tilde{z}_j, \tilde{y}_j)\}_{j=1}^{M}$ is obtained, which is then combined with the original embeddings to form the augmented dataset $\mathcal{Z}_{\text{aug}} = \mathcal{Z} \cup \mathcal{Z}_{\text{synthetic}}$. 
This approach increases sample diversity without requiring synthesis of raw signals or spectrograms. The mapping $g_{\text{LLM}}(\cdot)$ is realized through a structured prompting strategy, where the context set $S$ is serialized into a conditioning prefix $C$ and provided to the LLM. As illustrated in Appendix~\ref{appx:prompt} and Fig.~\ref{fig:prompts-generator}, the prompt consists of a system-level instruction and a user-level input. The system prompt defines the generation objective and enforces constraints such as preserving class structure, maintaining proximity to the underlying embedding manifold, and introducing controlled variability without producing outliers.

The user prompt provides the few-shot context set $S$ and explicitly requests the generation of new embeddings that follow the same distribution. The language model then models the joint probability distribution of the novel embedding tokens conditioned on this prefix, evaluating the conditional probability $P(\tilde{z} | S)$. Because the model has internalized extensive numerical and structural patterns during its initial training, this structured prompt enables the LLM to infer the geometric relationships within the embedding space through in-context learning. By conditioning on a small number of labeled examples, the model effectively acts as a conditional generator, restricting its output to the local feature manifold defined by the real examples. Importantly, this approach avoids the need to train a dedicated generative model. By combining a pretrained feature extractor with in-context generation, this framework provides a lightweight, efficient mechanism to increase sample diversity without ever requiring the synthesis of raw signals or spectrograms. 

The following section describes the experimental setup used to evaluate the proposed \llm framework.

\section{Experimental Setup}
\label{sec:experimental_setup}
In this section, we evaluate \llm from several perspectives. We examine data efficiency by measuring the amount of labeled data required to approach full-data performance. We compare \llm against no augmentation and generative baselines under low-shot settings with $D \in {10, 25, 50}$ samples per class. Experiments are conducted on two datasets: RadioML \cite{oshea2016radio} for modulation classification and the interference classification (IC) dataset \cite{chiejina2024system}. We also study the effect of different augmentation factors, including $1.2\times$, $1.5\times$, and $2\times$, where the factor denotes the ratio of the augmented training set size to the original dataset. We leverage a stratified $80/20$ train-test split and reserve 10\% of the training data for validation, again via stratified sampling, for early stopping. The same data-splitting criterion is used for both datasets.

\subsection{Datasets}

\noindent \textbf{RadioML \cite{oshea2016radio}:} We evaluate our approach on the RadioML 2016.10A dataset, a widely used benchmark for modulation classification in wireless learning. The dataset consists of time-frequency spectrogram representations of radio signals, which are used as inputs to the learning pipeline. Each sample is represented as a three-channel spectrogram of size $(3, 64, 64)$, corresponding to different signal components. Following standard practice, we focus on the SNR $=18$ dB slice and consider all 11 modulation classes. 


\noindent \textbf{Interference Classification (IC) Dataset \cite{chiejina2024system}:} In addition to RadioML, we evaluate \llm on an interference classification (IC) dataset derived from an O-RAN xApp deployment. The dataset consists of three classes corresponding to different interference conditions: signal-of-interest (SOI), SOI with co-channel interference (SOI+CI), and SOI with continuous wave interference (SOI+CWI). A subset of 700 samples per class was leveraged in \cite{gajjar2024preserving}, resulting in a total of 2,100 spectrogram images. All samples are resized to $64 \times 64$ RGB representations to maintain consistency with the RadioML input format. 


\subsection{Baselines}

\noindent \textbf{Baseline Methods:} We compare \llm against a set of standard augmentation baselines commonly used in wireless learning. As a reference point, we include a \textit{no-augmentation} setting, where the classifier is trained using only $D$ real samples per class. 
We then consider three representative generative approaches: \textit{conditional Generative Adversarial Networks (\textbf{GAN})}, \textit{conditional Variational Autoencoders (\textbf{VAE})}, and \textit{diffusion-based models (\textbf{DDPM})}\footnote{Detailed background for the baseline approaches is available in Appendix C}. For each method, synthetic samples are generated in a class-conditional manner and combined with the real samples, resulting in a total of $2D$ training samples per class. All baselines follow the same augmentation protocol and training pipeline to ensure a fair comparison in low-shot settings. All experiments use a common simple convolutional neural network (SimpleCNN) architecture as the feature extractor and classification backbone. The network consists of two convolutional layers with ReLU activations and max-pooling, followed by a fully connected layer that produces a 128-dimensional feature representation, and a final classification layer. This shared architecture ensures that performance differences across methods are attributable to the augmentation strategy rather than model capacity. The network is trained using the Adam optimizer with early stopping based on validation performance. We further designate \textit{DDPM} as our primary generative baseline. As discussed in Section \ref{sec:background}, recent literature firmly establishes diffusion models as the state-of-the-art approach for data augmentation in machine learning.

\subsection{Experimental Configuration}



\noindent \textbf{Feature Extractor $f_{\theta}(\cdot)$:} The feature extractor is implemented using the SimpleCNN architecture described earlier and is trained in the same manner as the no-augmentation baseline. It maps input spectrograms to 128-dimensional embedding vectors that capture class-relevant structure. This design decouples representation learning from the augmentation process, allowing the subsequent stages to operate entirely in the embedding space without requiring task-specific generative model training. 


\noindent \textbf{Large Language Model $g_{\text{LLM}}$:} The LLM operates on a small set of labeled embedding vectors provided through in-context prompts. We vary the number of prompt examples per class, denoted by $k$, over the set $\{5, 10, 15\}$, with $k=5$ used when $D=10$ due to limited sample availability. For each configuration, we report the best-performing results, while a detailed analysis of the impact of $k$ is provided in Appendix~\ref{app:optimal_k}.


\noindent \textbf{Classifier $MLP$:} The final classifier is a multilayer perceptron (MLP) trained on the combined set of real and synthetic embeddings after projection using Linear Discriminant Analysis (LDA). The projection reduces the embedding dimensionality to $n_{\text{comp}} = \text{num\_classes} - 1$, improving class separability while maintaining compact representations. The MLP consists of a single hidden layer with ReLU activation, followed by an output layer for classification.

\subsection{Evaluation Metrics}
To evaluate classification models, we use a combination of metrics based on the confusion matrix components\ True Positives ($TP$), True Negatives ($TN$), False Positives ($FP$), and False Negatives ($FN$) which are calculated as follows: $Accuracy = \frac{TP + TN}{TP + TN + FP + FN}$, $Precision = \frac{TP}{TP + FP}$, $Recall = \frac{TP}{TP + FN}$, and $F1 = 2 \times \frac{Precision \times Recall}{Precision + Recall}$.

\section{Results}
\subsection{Data Efficiency}

\begin{figure}[h]
    \centering
    \includegraphics[width=0.9\linewidth]{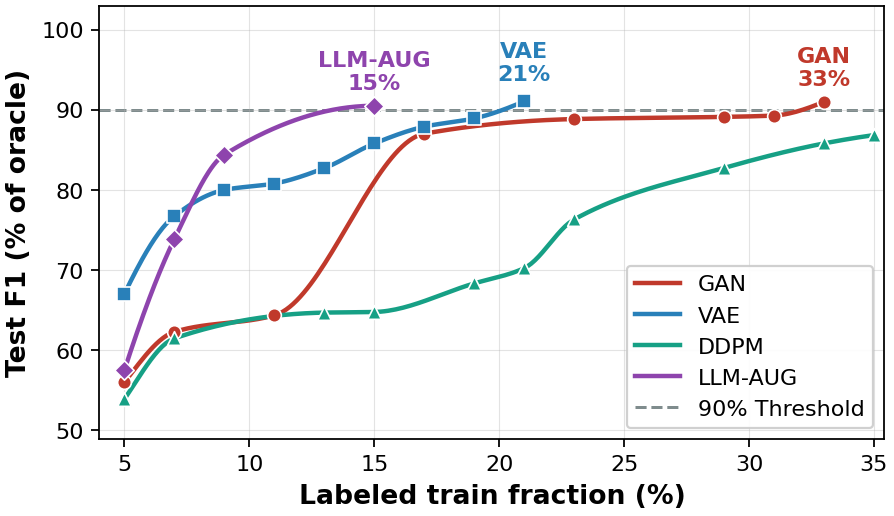}
    \caption{Normalized test F$_1$ vs.\ labeled data. \llm reaches 90\% oracle performance with fewer samples than generative baselines.}
    \label{fig:data_efficiency}
    
\end{figure}

Data efficiency is measured as the amount of labeled data required to approach the performance of an oracle model trained on the full training set. Starting from small fractions of the available training data, we progressively increase the number of labeled samples and compare \llm against generative baselines, including GAN, VAE, and diffusion models, under identical sampling conditions. Fig.~\ref{fig:data_efficiency} plots test F$_1$ (normalized to the full-data oracle) as a function of labeled training fraction and shows how quickly each method approaches oracle-level performance as the amount of labeled data increases.
\llm consistently achieves higher performance at lower data regimes and reaches the 90\% oracle threshold with substantially fewer labeled samples than all generative baselines. Specifically, \llm reaches this threshold at approximately 15\% labeled data, whereas VAE requires around 21\% and GAN approximately 33\%. DDPM does not reach this performance level within the evaluated range.
This gap highlights a fundamental difference in how synthetic samples are generated. While GAN, VAE, and DDPM attempt to model high-dimensional spectrogram distributions from limited data, \llm operates in a structured embedding space derived from CNN features and LDA projections, allowing it to produce more informative and class-consistent augmentations even in extremely low-data regimes. Notably, all methods follow the same $2\times$ augmentation protocol, ensuring that improvements arise from sample quality rather than increased data volume.

Overall, these results demonstrate that \llm is significantly more data-efficient, achieving near-oracle performance with substantially fewer labeled examples. 

These trends are consistent with the discrete low-shot results in Table~\ref{tab:modulation_performance_wide}. At $D=10$, \llm achieves an F$_1$ score of 36.2\%, outperforming both the no-augmentation baseline (30.1\%) and all generative methods, including GAN (29.3\%), VAE (24.9\%), and DDPM (21.6\%), translating to a relative gain of 23.5\% over GAN. Similar gains are observed at $D=25$, where \llm attains the highest performance among all methods. At $D=50$, performance differences become smaller, with generative baselines improving as more data becomes available, while \llm remains competitive. These results reinforce that the largest gains occur in the most data-constrained regimes, consistent with the data efficiency trends shown in Fig.~\ref{fig:data_efficiency}.

\subsection{Stress Tests}
\begin{table*}[t]
\centering
\scriptsize
\caption{Stress tests on the RadioML dataset}
\label{tab:modulation_performance_wide}
\resizebox{\textwidth}{!}{
\begin{tabular}{l|cccc|cccc|cccc}
\hline
\multirow{2}{*}{\textbf{Method}} & \multicolumn{4}{c|}{\textbf{10 s/cls}} & \multicolumn{4}{c|}{\textbf{25 s/cls}} & \multicolumn{4}{c}{\textbf{50 s/cls}} \\
\cline{2-13}
 & \textbf{F1} & \textbf{Acc} & \textbf{Prec} & \textbf{Rec} & \textbf{F1} & \textbf{Acc} & \textbf{Prec} & \textbf{Rec} & \textbf{F1} & \textbf{Acc} & \textbf{Prec} & \textbf{Rec} \\
\hline
No Augmentation & 30.1\% & 35.2\% & 31.8\% & 35.2\% & 44.1\% & 45.1\% & 44.1\% & 45.1\% & 40.7\% & 44.5\% & 42.4\% & 44.5\% \\
\hline
\multicolumn{13}{c}{\textbf{2x Experiments}} \\
\hline
GAN & 29.3\% & 30.0\% & 33.0\% & 30.0\% & 41.3\% & 42.2\% & 45.5\% & 42.2\% & 44.6\% & 47.3\% & 47.0\% & 47.3\% \\
VAE & 24.9\% & 28.1\% & 26.5\% & 28.1\% & 38.1\% & 40.3\% & 40.2\% & 40.3\% & 47.9\% & \textbf{49.7\%} & 48.8\% & \textbf{49.7\%} \\
Diffusion & 21.6\% & 29.9\% & 27.7\% & 29.9\% & 36.2\% & 38.6\% & 38.6\% & 38.6\% & 44.3\% & 46.4\% & 44.1\% & 46.4\% \\
\textbf{LLM-AUG (Ours)} & \textbf{36.2\%} & \textbf{38.0\%} & \textbf{41.6\%} & \textbf{38.0\%} & \textbf{48.3\%} & \textbf{48.5\%} & \textbf{48.5\%} & \textbf{48.5\%} & \textbf{48.9\%} & 49.3\% & \textbf{48.8\%} & 49.3\% \\
\hline
\multicolumn{13}{c}{\textbf{1.5x Experiments}} \\
\hline
GAN & 31.5\% & 34.0\% & 31.6\% & 34.0\% & 41.2\% & 43.0\% & 45.4\% & 43.0\% & 43.4\% & 45.8\% & 46.1\% & 45.8\% \\
VAE & 21.9\% & 31.3\% & 25.0\% & 31.3\% & 41.5\% & 42.8\% & 43.5\% & 42.8\% & \textbf{53.6\%} & \textbf{54.0\%} & \textbf{54.6\%} & \textbf{54.0\%} \\
Diffusion & 14.3\% & 26.8\% & 14.6\% & 26.8\% & 19.2\% & 27.1\% & 28.8\% & 27.1\% & 46.0\% & 47.4\% & 45.6\% & 47.4\% \\
\textbf{LLM-AUG (Ours)} & \textbf{36.3\%} & \textbf{37.7\%} & \textbf{43.3\%} & \textbf{37.7\%} & \textbf{44.8\%} & \textbf{45.7\%} & \textbf{45.6\%} & \textbf{45.7\%} & 48.5\% & 48.9\% & 48.3\% & 48.9\% \\
\hline
\multicolumn{13}{c}{\textbf{1.2x Experiments}} \\
\hline
GAN & 29.2\% & 34.0\% & 30.9\% & 34.0\% & 43.7\% & 45.0\% & 43.5\% & 45.0\% & \textbf{54.9\%} & \textbf{55.5\%} & \textbf{56.2\%} & \textbf{55.5\%} \\
VAE & 27.4\% & 33.6\% & 28.4\% & 33.6\% & 28.5\% & 33.5\% & 31.4\% & 33.5\% & 46.8\% & 49.1\% & 49.8\% & 49.1\% \\
Diffusion & 29.6\% & 33.7\% & 30.9\% & 33.7\% & 35.3\% & 37.3\% & 38.4\% & 37.3\% & 41.2\% & 46.6\% & 43.4\% & 46.6\% \\
\textbf{LLM-AUG (Ours)} & \textbf{37.0\%} & \textbf{38.3\%} & \textbf{44.6\%} & \textbf{38.3\%} & \textbf{46.1\%} & \textbf{47.3\%} & \textbf{46.9\%} & \textbf{47.3\%} & 47.5\% & 48.7\% & 47.3\% & 48.7\% \\
\hline
Full-Data Oracle & 75.4\% & 76.1\% & 76.7\% & 76.1\% & 75.4\% & 76.1\% & 76.7\% & 76.1\% & 75.4\% & 76.1\% & 76.7\% & 76.1\% \\
\hline
\end{tabular}%
}

\end{table*}

We next evaluate performance under extreme low-shot conditions to isolate the relative performance improvements provided by each augmentation method. As shown in Table~\ref{tab:modulation_performance_wide}, generative approaches exhibit significant degradation when labeled data is highly limited. In particular, DDPM performs poorly at $D=10$, achieving an F$_1$ score of 21.6\%, well below both the no-augmentation baseline and other methods. VAE and GAN also show inconsistent behavior across low-shot regimes, indicating difficulty in learning reliable class-conditional distributions from limited data. In contrast, \llm maintains stable and consistent improvements across all settings. At $D=10$, it achieves 36.2\% F$_1$, significantly outperforming all baselines, and continues to provide competitive or superior performance at $D=25$ and $D=50$. These results demonstrate that \llm is more robust in data-scarce regimes, where generative models are prone to degradation or failure.



\subsection{Hard Classes:} 
\begin{figure}
    \centering
    \includegraphics[width=0.9\linewidth]{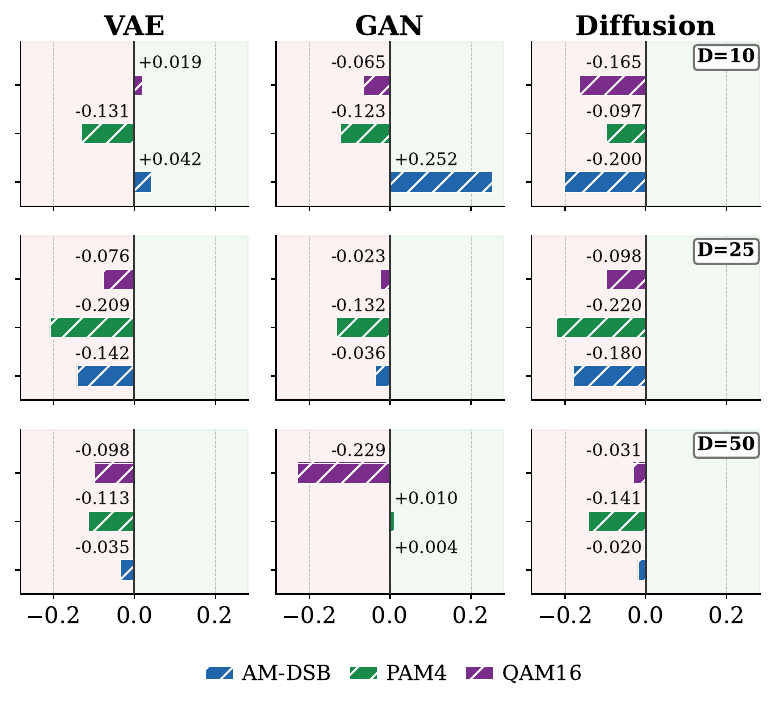}
    \caption{Class-wise $\Delta$F$_1$ relative to LLM-AUG across VAE, GAN, and diffusion for $D=10,25,50$.}
    \label{fig:hard_classes}
    
\end{figure}

We further analyze class-level behavior by identifying the hardest classes to improve under augmentation. For each method and data regime, we compute $\Delta \text{F1} = \text{F1}(\text{method}) - \text{F1}(\text{none})$, and aggregate across all $(D, \text{method})$ pairs. Based on the mean $\Delta \text{F1}$, the three most challenging classes are AM-DSB ($-0.072$), PAM4 ($-0.069$), and QAM16 ($-0.028$). As shown in Fig.~\ref{fig:hard_classes}, all methods struggle on these classes, exhibiting negative average gains relative to the no-augmentation baseline. For AM-DSB and PAM4, this degradation is consistent across all methods, including the proposed \texttt{LLM-AUG} framework, indicating that these classes remain difficult to improve under low-shot augmentation regardless of the generation strategy. In contrast, for QAM16, \llm demonstrates improved performance over generative baselines despite the overall difficulty, suggesting that operating in the embedding space allows it to better preserve class structure when augmentation gains are limited.

\subsection{Robustness:} 
We evaluate robustness under distributional shift by testing models trained at 18~dB SNR on lower SNR conditions. As shown in Table~\ref{tab:snr_robustness_results}, \llm consistently outperforms DDPM across most settings, particularly at higher low-shot regimes. At $D=25$, \llm achieves up to 49.3\% F$_1$ at 16~dB, compared to 38.1\% for DDPM, indicating a substantial performance gap under mismatched conditions. More specifically, \llm achieves an improvement of 29.4\% over diffusion-based method. While both methods struggle at $D=10$ and lower SNR values, \llm maintains a consistent advantage. This suggests that the representations learned through \llm are more stable under shifts in signal conditions, whereas diffusion-based augmentation appears more sensitive to distribution mismatch. The consistent gap across SNR levels indicates that improvements from \llm are not limited to the training distribution, but extend to unseen operating regimes.

\begin{table*}[t]
\centering
\scriptsize
\caption{Evaluating Robustness}
\label{tab:snr_robustness_results}
\resizebox{\textwidth}{!}{%
\begin{tabular}{l|cccc|cccc|cccc|cccc}
\hline
\multirow{2}{*}{\textbf{Method}} 
& \multicolumn{4}{c|}{\textbf{SNR 10}} 
& \multicolumn{4}{c|}{\textbf{SNR 12}} 
& \multicolumn{4}{c|}{\textbf{SNR 14}} 
& \multicolumn{4}{c}{\textbf{SNR 16}} \\
\cline{2-17}
& \textbf{F1} & \textbf{Acc} & \textbf{Prec} & \textbf{Rec}
& \textbf{F1} & \textbf{Acc} & \textbf{Prec} & \textbf{Rec}
& \textbf{F1} & \textbf{Acc} & \textbf{Prec} & \textbf{Rec}
& \textbf{F1} & \textbf{Acc} & \textbf{Prec} & \textbf{Rec} \\
\hline

\multicolumn{17}{c}{\textbf{10 samples per class}} \\
\hline
Diffusion & 22.1\% & 29.8\% & 26.9\% & 29.8\% & 22.9\% & 30.6\% & 30.8\% & 30.6\% & 22.2\% & 30.1\% & 29.8\% & 30.1\% & 21.3\% & 29.6\% & 33.0\% & 29.6\% \\
\textbf{LLM-AUG (Ours)} & 26.3\% & 30.0\% & 33.3\% & 30.0\% & 33.9\% & 37.1\% & 37.8\% & 37.1\% & 36.8\% & 39.7\% & 39.4\% & 39.7\% & 38.7\% & 40.4\% & 42.2\% & 40.4\% \\

\hline
\multicolumn{17}{c}{\textbf{25 samples per class}} \\
\hline
Diffusion & 33.7\% & 36.6\% & 36.7\% & 36.6\% & 35.6\% & 37.7\% & 39.1\% & 37.7\% & 36.4\% & 38.8\% & 39.4\% & 38.8\% & 38.1\% & 39.8\% & 41.1\% & 39.8\% \\
\textbf{LLM-AUG (Ours)} & 31.4\% & 35.1\% & 37.1\% & 35.1\% & 41.7\% & 44.0\% & 45.8\% & 44.0\% & 47.3\% & 48.0\% & 48.6\% & 48.0\% & 49.3\% & 49.5\% & 49.7\% & 49.5\% \\

\hline
\multicolumn{17}{c}{\textbf{50 samples per class}} \\
\hline
Diffusion & 42.8\% & 44.8\% & 43.2\% & 44.8\% & 42.2\% & 44.2\% & 43.0\% & 44.2\% & 41.8\% & 44.0\% & 41.7\% & 44.0\% & 42.9\% & 45.1\% & 42.1\% & 45.1\% \\
\textbf{LLM-AUG (Ours)} & 43.0\% & 44.7\% & 44.9\% & 44.7\% & 47.5\% & 48.5\% & 48.0\% & 48.5\% & 47.7\% & 48.6\% & 47.5\% & 48.6\% & 47.9\% & 48.2\% & 47.7\% & 48.2\% \\

\hline
\end{tabular}%
}

\end{table*}

\subsection{t-SNE Visualizations:} 
\begin{figure}[t]
    \centering
    \includegraphics[width=\linewidth]{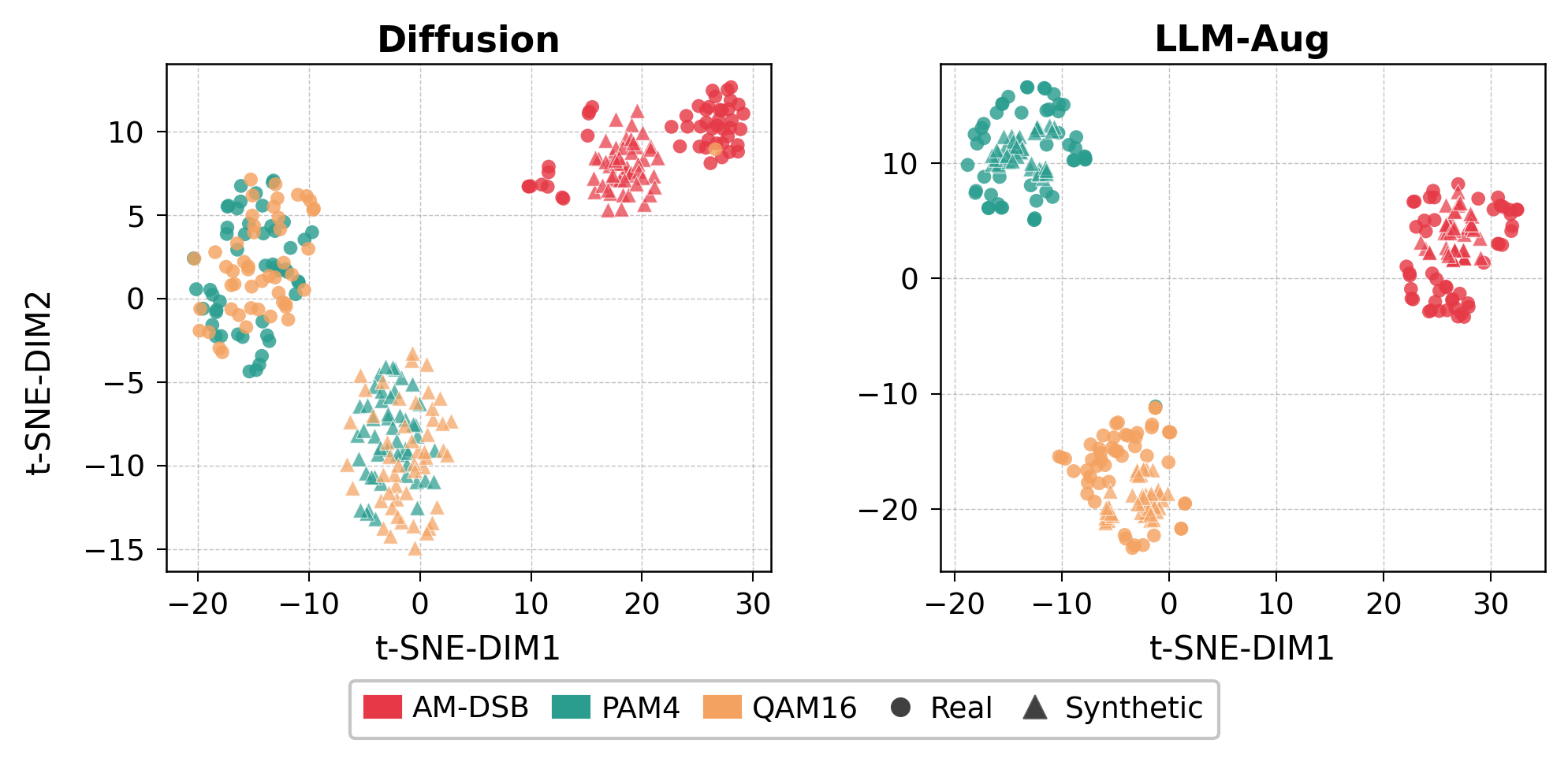}
    \caption{t-SNE projections of embedding space for diffusion (left) and LLM-AUG (right) at $D=50$, $k=15$. LLM-AUG shows more compact, class-aligned clusters.}
    \label{fig:tsne}
\end{figure}

We provide a qualitative comparison of the learned representations using t-SNE projections of the embedding space. As shown in Fig.~\ref{fig:tsne}, synthetic samples generated by \llm closely align with the corresponding real sample clusters, preserving class-wise structure in the reduced space. In contrast, synthetic samples from DDPM appear more diffuse and less aligned with the real data distribution, indicating weaker class separation. These observations suggest that \llm produces more structured and class-consistent augmentations in the embedding space, while diffusion-based augmentation struggles to capture similar structure under low-shot conditions.

\begin{table*}[t]
\centering
\scriptsize
\caption{Stress tests on the IC dataset} 
\label{tab:icxapp_wide}
\resizebox{\textwidth}{!}{%
\begin{tabular}{l|cccc|cccc|cccc}
\hline
\multirow{2}{*}{\textbf{Method}} & \multicolumn{4}{c|}{\textbf{10 s/cls}} & \multicolumn{4}{c|}{\textbf{25 s/cls}} & \multicolumn{4}{c}{\textbf{50 s/cls}} \\
\cline{2-13}
 & \textbf{F1} & \textbf{Acc} & \textbf{Prec} & \textbf{Rec} & \textbf{F1} & \textbf{Acc} & \textbf{Prec} & \textbf{Rec} & \textbf{F1} & \textbf{Acc} & \textbf{Prec} & \textbf{Rec} \\
\hline
No Augmentation & 57.2\% & 57.6\% & 58.0\% & 57.6\% & 30.5\% & 38.1\% & 25.4\% & 38.1\% & 55.5\% & 58.3\% & 58.5\% & 58.3\% \\
\hline
\multicolumn{13}{c}{\textbf{2x Experiments}} \\
\hline
GAN & 48.5\% & 52.4\% & 52.2\% & 52.4\% & 25.9\% & 27.1\% & 26.8\% & 27.1\% & 70.8\% & 71.4\% & 74.1\% & 71.4\% \\
VAE & \textbf{55.4\%} & \textbf{57.6\%} & \textbf{68.2\%} & \textbf{57.6\%} & 65.9\% & 67.1\% & 67.7\% & 67.1\% & 81.6\% & 82.1\% & 82.0\% & 82.1\% \\
Diffusion & 16.7\% & 33.3\% & 11.1\% & 33.3\% & 52.4\% & 52.9\% & 52.6\% & 52.9\% & 76.2\% & 76.9\% & 77.0\% & 76.9\% \\
\textbf{LLM-AUG (Ours)} & 52.7\% & 57.4\% & 66.3\% & 57.4\% & \textbf{71.1\%} & \textbf{70.7\%} & \textbf{76.4\%} & \textbf{70.7\%} & \textbf{84.8\%} & \textbf{85.0\%} & \textbf{84.8\%} & \textbf{85.0\%} \\
\hline
\multicolumn{13}{c}{\textbf{1.5x Experiments}} \\
\hline
GAN & 16.7\% & 33.3\% & 11.1\% & 33.3\% & 39.3\% & 44.8\% & 48.8\% & 44.8\% & 73.7\% & 74.8\% & 76.1\% & 74.8\% \\
VAE & \textbf{52.6\%} & 51.9\% & 54.8\% & 51.9\% & 59.0\% & 59.3\% & 59.5\% & 59.3\% & 77.2\% & 78.6\% & 80.4\% & 78.6\% \\
Diffusion & 16.7\% & 33.3\% & 11.1\% & 33.3\% & 58.2\% & 59.8\% & 65.7\% & 59.8\% & 59.0\% & 58.8\% & 62.1\% & 58.8\% \\
\textbf{LLM-AUG (Ours)} & 49.1\% & \textbf{52.6\%} & \textbf{63.8\%} & \textbf{52.6\%} & \textbf{79.9\%} & \textbf{81.2\%} & \textbf{82.8\%} & \textbf{81.2\%} & \textbf{79.5\%} & \textbf{80.2\%} & \textbf{81.2\%} & \textbf{80.2\%} \\
\hline
\multicolumn{13}{c}{\textbf{1.2x Experiments}} \\
\hline
GAN & 16.7\% & 33.3\% & 11.1\% & 33.3\% & 37.6\% & 47.6\% & 33.4\% & 47.6\% & 77.1\% & 79.3\% & 83.7\% & 79.3\% \\
VAE & \textbf{55.7\%} & \textbf{56.0\%} & 56.6\% & \textbf{56.0\%} & 64.1\% & 64.0\% & 64.3\% & 64.0\% & 68.7\% & 69.8\% & 68.7\% & 69.8\% \\
Diffusion & 39.5\% & 49.8\% & 34.5\% & 49.8\% & 36.0\% & 45.0\% & 30.0\% & 45.0\% & 69.9\% & 70.5\% & 69.8\% & 70.5\% \\
\textbf{LLM-AUG (Ours)} & 49.9\% & 53.1\% & \textbf{64.4\%} & 53.1\% & \textbf{74.9\%} & \textbf{77.4\%} & \textbf{81.3\%} & \textbf{77.4\%} & \textbf{83.9\%} & \textbf{84.5\%} & \textbf{85.5\%} & \textbf{84.5\%} \\
\hline
Full-Data Oracle & 96.4\% & 96.4\% & 96.6\% & 96.4\% & 96.4\% & 96.4\% & 96.6\% & 96.4\% & 96.4\% & 96.4\% & 96.6\% & 96.4\% \\
\hline
\end{tabular}%
}
\end{table*}
\subsection{Extended Evaluation with IC Dataset:}

We further evaluate the proposed approach on the Interference Classification (IC) dataset to assess generalization across tasks. As shown in Table~\ref{tab:icxapp_wide}, \llm achieves the best performance at $D=25$ and $D=50$ across all augmentation ratios ($2\times$, $1.5\times$, $1.2\times$). In particular, at $D=50$ under the $2\times$ setting, \llm attains 84.8\% F$_1$, outperforming VAE (81.6\%), DDPM (76.2\%), and GAN (70.8\%), translating to 19.8\% improvement over GAN. At lower data regimes, performance is more sensitive to the specific samples selected for training. Since the dataset is relatively simpler, small subsets (e.g., $D=10$ or $D=25$) can be either highly representative or less informative, leading to variability in observed performance. Despite this, \llm remains competitive across all settings and consistently outperforms generative baselines in moderate to higher data regimes.

These results indicate that the relative advantages of \llm extend beyond RadioML, while also highlighting that the impact of augmentation depends on dataset complexity and sample representativeness in low-shot scenarios.

\section{Discussion and Limitations}

While our work showcases significant advancements for robust data augmentation in wireless signal classification tasks, we acknowledge the following limitations and open challenges: 

\noindent $\bullet$ \textbf{Hyperparameter Selection and Token Constraints:} Our empirical evaluation heavily relies on specific choices for the hyperparameters $\alpha$ and $\beta$, which were intentionally restricted due to token limitations and their associated costs. Conducting extensive experiments across a wider range of these variables would increase token consumption exponentially. While we believe our selected values adequately capture and explain the target hyperspace for this study, future work should more rigorously explore the $\alpha$ and $\beta$ parameter space to fully understand their impact on model performance.

\noindent $\bullet$ \textbf{Feature Extractor Architecture:} Throughout this study, we deliberately maintained a lightweight feature extractor design to prioritize computational efficiency and rapid experimentation. However, there is significant potential in transitioning towards a more sophisticated architectural setup. Future research could greatly benefit from integrating established, complex pretrained models, such as ResNet or MobileNet, to potentially yield richer feature representations and improve overall robustness.

\noindent $\bullet$ \textbf{Modeling Pipeline Alternatives:} Our current methodology employs Linear Discriminant Analysis (LDA) as the primary technique within our pipeline. While LDA serves our current objectives effectively, we acknowledge that there are numerous alternative dimensionality reduction and classification techniques available in the literature \cite{sorzano2014surveydimensionalityreductiontechniques}. An exhaustive evaluation and comparison of these alternative methods could provide deeper insights, but it remains strictly outside the current scope of this study.

\noindent $\bullet$ \textbf{Foundation Model Constraints:} 
The current framework leverages a single large language model, Gemma3-27B \cite{gemmateam2025gemma3technicalreport}, to conduct all augmentation tasks. Although exploring other reasoning-capable LLMs as a foundational base could provide broader insights and validate the generalizability of our approach, such comparative studies were precluded by severe computational constraints. Expanding to other open weight models and better paid models like GPT-5.2 \cite{singh2025openaigpt5card, _2025_introducing} is a primary objective for future iterations of this work. As it is intuitive to believe that leveraging a model with better reasoning capabilities should yield better synthetic samples.

\noindent $\bullet$ \textbf{Context Window and Scaling $k$:} 
As observed in our ablation study in Appendix \ref{app:optimal_k}, we hypothesize that employing higher values for the parameter $k$ might yield better overall accuracy by providing the model with more comprehensive examples. Nevertheless, scaling this parameter inherently increases the risk of the models becoming susceptible to context overload \cite{liu2024lost}, which can paradoxically degrade their reasoning capabilities. Balancing the value of $k$ against the limitations of the model's context window remains a critical challenge to be addressed in subsequent research.



\section{Conclusion}

This paper introduced \texttt{LLM-AUG}, an embedding-space data augmentation framework that leverages the in-context learning capabilities of large language models to address data scarcity in wireless communications. By operating in a compact feature space, the proposed approach avoids the complexities of generating raw RF signals while enabling efficient, scalable augmentation without training task-specific generative models. Through extensive evaluation across both modulation and interference classification tasks, we demonstrated that \texttt{LLM-AUG} provides strong, consistent improvements over classical deep generative methods in low-shot regimes. Specifically, the framework achieves near-Oracle performance using only \textasciitilde 15\% of the labeled data, and when applied on the RadioML and IC datasets, outperforms the GAN baseline by about 23.5\% and 19.8\%, respectively. And when applied on the RadioML and IC datasets, \texttt{LLM-AUG} framework outperforms DDPM by about 67.6\% and 35.7\%, respectively. These quantitative gains are supported by t-SNE visualizations, which confirm that our synthetic samples better preserve class structure within the embedding space, leading to highly informative augmentations. Furthermore, \texttt{LLM-AUG} maintains remarkable robustness under distribution shifts, yielding around 29.4\% relative gain over DDPM when evaluated at lower, unseen SNR values. Beyond the specific tasks considered here, \texttt{LLM-AUG} provides a general, highly adaptable framework for leveraging foundation models in structured signal domains. Future work will explore mitigating the aforementioned limitations and scaling to higher-dimensional embedding spaces, incorporating more advanced feature extractors, and extending the framework to additional wireless tasks and real-world deployments.


\bibliography{res.bib}
\bibliographystyle{ieeetr}
\appendix
\subsection{Model and Concept Drift}\label{appx:drift}

The simulation models two phenomena across three time steps ($t_0$, $t_1$, $t_2$), with all data sampled from bivariate normal distributions. For \textit{Model Drift}, Two classes, A and B, maintain static covariance matrices $\Sigma_A$ and $\Sigma_B$ while their means $\mu_A(t)$ and $\mu_B(t)$ migrate over time. A linear decision boundary is fixed at $t_0$, passing through the midpoint $M = \frac{\mu_A(t_0) + \mu_B(t_0)}{2}$ and oriented perpendicular to the inter-class axis. Given displacement $V = (V_x, V_y)$ where $\mu_B(t_0) = \mu_A(t_0) + V$, the boundary slope $m$ satisfies:
\begin{equation}
mV_y + V_x = 0, \qquad c = M_y - mM_x
\end{equation}
As the distributions translate while the boundary remains fixed, classification performance degrades.   For \textit{Concept Drift}, three classes evolve with both means $\mu_C(t)$ and covariances $\Sigma_C(t)$ changing at each time step, capturing shifts in cluster shape, variance, and feature correlations. At $t_2$, the third class transitions into a Gaussian mixture:
\begin{equation}
P(x) = w_1\,\mathcal{N}(x \mid \mu_\text{new}, \Sigma_\text{new}) + w_2\,\mathcal{N}(x \mid \mu_\text{old}, \Sigma_\text{old})
\end{equation}
where $\quad w_1 + w_2 = 1$ . This bifurcation models the emergence of a novel behavioral mode within an existing category.

\subsection{Token Costs}\label{appx:multimodality}

The cost ratios depicted in Figure \ref{fig:cost} illustrate the fundamental scaling differences between processing high-dimensional visual data and compact numerical vectors. Generally, commercial vision language models process images by dividing them into fixed pixel patches using a Vision Transformer encoder. Because each patch is converted into a separate token, the total token count inherently scales quadratically with the image resolution. Furthermore, commercial application programming interfaces typically impose premium pricing for visual tokens due to the increased computational overhead of image processing and generation. To calculate the specific ratios in Figure \ref{fig:cost}, we utilized the pricing structure of the current OpenAI suite of models as a representative example. Under this commercial tier, image input tokens derived from $16 \times 16$ pixel patches are billed at \$8.00 per million tokens, whereas standard text tokens utilized for our 10-dimensional embedding baseline cost only \$1.75 per million. 

On the generation side, autoregressive image outputs relying on finer $8 \times 8$ patches are billed at a premium of \$32.00 per million tokens, compared to the text generation baseline of \$14.00 per million. The combined effect of this higher per-token rate and the quadratically scaling token count drives the exponential cost multipliers observed when attempting to augment data directly in the visual domain.
\subsection{Prompt Template}\label{appx:prompt}
\begin{figure}[]
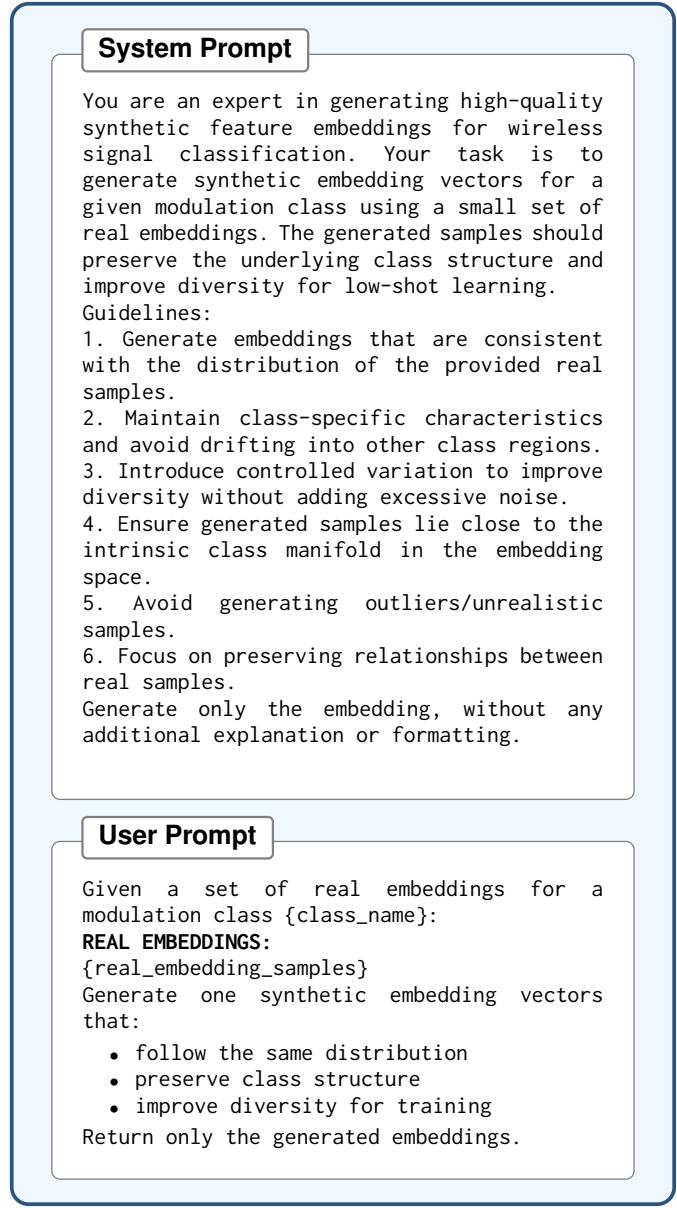

\begin{minipage}[]{\linewidth}
\begin{tcolorbox}[
  colback=genbg,
  colframe=genframe,
  drop shadow,
  arc=2mm
]
\begin{promptbox}{System Prompt}
You are an expert in generating high-quality synthetic feature embeddings for wireless signal classification.
Your task is to generate synthetic embedding vectors for a given modulation class using a small set of real embeddings. The generated samples should preserve the underlying class structure and improve diversity for low-shot learning.\\
Guidelines:\\
1. Generate embeddings that are consistent with the distribution of the provided real samples.\\
2. Maintain class-specific characteristics and avoid drifting into other class regions.\\
3. Introduce controlled variation to improve diversity without adding excessive noise.\\
4. Ensure generated samples lie close to the intrinsic class manifold in the embedding space.\\
5. Avoid generating outliers/unrealistic samples.\\
6. Focus on preserving relationships between real samples.\\
Generate only the embedding, without any additional explanation or formatting.\\
\end{promptbox}

\begin{promptbox}{User Prompt}
Given a set of real embeddings for a modulation class \texttt{\{class\_name\}}:

\textbf{REAL EMBEDDINGS:}

\texttt{\{real\_embedding\_samples\}}

Generate \textit{one} synthetic embedding vectors that:
\begin{itemize}
    \item follow the same distribution
    \item preserve class structure
    \item improve diversity for training
\end{itemize}

Return only the generated embeddings.
\end{promptbox}
\end{tcolorbox}
\caption{System and User Prompts for \texttt{LLM-AUG}.}\label{fig:prompts-generator}
\end{minipage}\hfill
\end{figure}

The prompt template for the system prompt and user prompt, as mentioned in Section \ref{subsec:llm_aug}, is showcased in Figure \ref{fig:prompts-generator}.

\subsection{Baseline Methods}

\subsubsection{No Augmentation}

The no-augmentation baseline serves as a reference point, where the classifier is trained using only $D$ real samples per class without any synthetic data generation. This setting reflects the performance achievable under strict low-data conditions and establishes a lower bound for comparison against augmentation-based methods. Since no additional data is introduced, the learned representations are entirely dependent on the limited training samples, often resulting in reduced generalization and increased sensitivity to noise and class overlap. All training configurations, including optimizer, architecture, and early stopping criteria are kept identical across all methods to ensure a fair and controlled comparison.

\subsubsection{GAN}

The GAN baseline is implemented using a conditional generative adversarial framework comprising a ConditionalGenerator and ConditionalDiscriminator. The generator maps a latent noise vector of dimension $z_{\text{dim}} = 100$, concatenated with class-conditioning information, to synthetic spectrogram samples, while the discriminator learns to distinguish real from generated samples under the same conditioning. This adversarial training setup enforces distributional alignment between synthetic and real data \cite{goodfellow2014generative}. Training is performed using the Adam optimizer with learning rate $2 \times 10^{-4}$ and $\beta=(0.5, 0.999)$, optimized with the BCE objective for 150 epochs and a batch size of 32. Following training, $D$ synthetic samples per class are generated and combined with real samples, yielding a total of $2D$ samples per class. While GANs can produce realistic samples, their performance in low-data regimes is limited by training instability and reduced sample diversity. In particular, mode collapse remains a key challenge, leading to insufficient coverage of intra-class variability, which is critical for discriminative tasks such as modulation classification. 

\subsubsection{VAE}

The VAE baseline uses a conditional variational autoencoder with a 4-layer encoder and decoder architecture. The encoder maps input spectrograms into a latent distribution characterized by a mean and variance, while the decoder reconstructs samples conditioned on both the latent variables and class labels. The model is trained using the evidence lower bound objective which balances reconstruction fidelity with latent space regularization \cite{kingma2013auto}.

Optimization is performed using the Adam optimizer with a learning rate of $2 \times 10^{-4}$ for 150 epochs and a batch size of 32. After training, $D$ synthetic samples per class are generated by sampling from the learned latent distribution and decoding them into spectrogram space yielding a total of $2D$ samples per class when combined with real data.

While VAEs provide stable training and structured latent representations they often produce overly smooth or blurred samples due to the variational constraint, which can limit their ability to capture sharp class boundaries and fine-grained signal characteristics required for modulation classification.

\subsubsection{DDPM}

The diffusion baseline is implemented using a MiniDDPM architecture with a U-Net backbone, which models the data distribution through a multi-step denoising process. The forward process gradually adds Gaussian noise to the data over $T=200$ steps, following a linear noise schedule with $\beta$ ranging from $1 \times 10^{-4}$ to $0.02$, while the reverse process learns to iteratively reconstruct clean samples from noisy inputs. The model is trained to predict the added noise at each timestep using a mean squared error objective \cite{ho2020denoising}. Training is performed using the Adam optimizer with a learning rate of $2 \times 10^{-4}$ for 150 epochs and a batch size of 32. During generation, the reverse diffusion process is used to synthesize $D$ samples per class, resulting in $2D$ total samples per class when combined with real data. Diffusion models are generally more stable than GANs and provide improved sample diversity; however, they are computationally expensive and may still struggle to fully capture class-specific structure in extremely low-data settings, where the learned denoising process lacks sufficient signal to model the underlying distribution accurately.

\subsection{Ablation Study: Optimal $k$}\label{app:optimal_k}
\begin{figure}[htbp]
    \centering
    \includegraphics[width=0.9\linewidth]{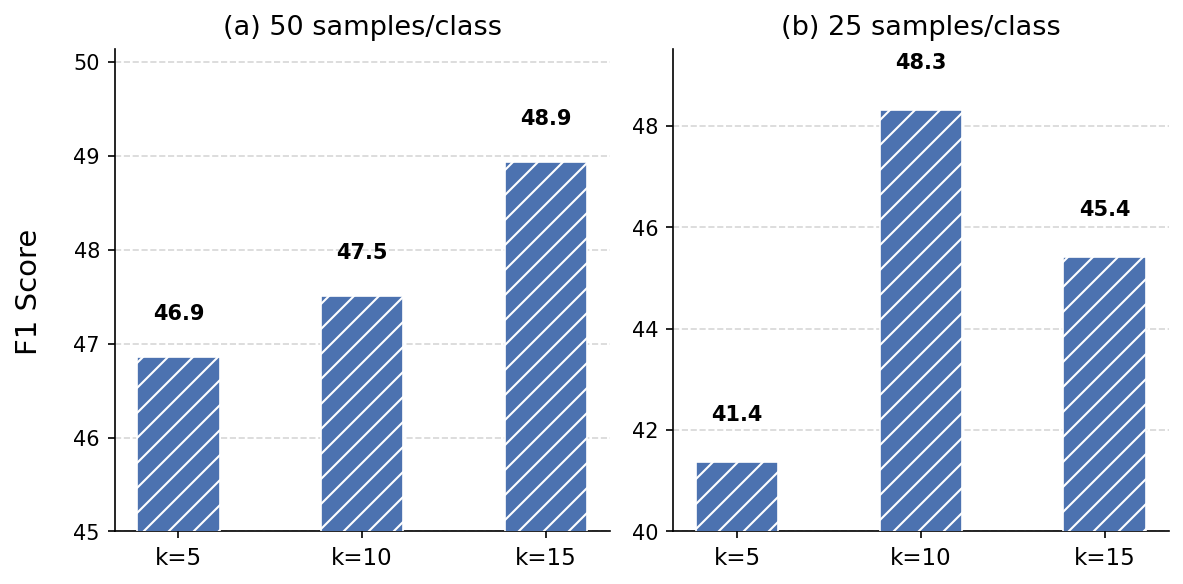}
    \caption{Ablation study evaluating the impact of the number of in-context examples ($k$) on classification accuracy. Results are shown for the 50 s/cls and 25 s/cls training sets.}
    \label{fig:k_ablation}
    
\end{figure}

To determine the optimal number of in-context demonstration examples ($k$), we conducted an ablation study on the RadioML dataset. We evaluated the downstream F1 score for $k \in \{5, 10, 15\}$, focusing primarily on a distribution of 50 samples per class, with a supplementary evaluation at 25 samples per class. As shown in Figure \ref{fig:k_ablation}, the F1 score consistently improves as $k$ increases in the 50 samples/class scenario, peaking at $k=15$ with a score of 48.9. However, in the 25 samples/class scenario, the F1 score peaks at $k=10$ before decreasing at $k=15$. While increasing $k$ further might offer the LLM a richer representation of the class distribution and potentially improve performance, it also risks exceeding effective context limits or causing context degradation \cite{liu2024lost}, which likely explains the performance drop observed at $k=15$ in the lower-sample regime. 
\end{document}